\documentclass{article} % For LaTeX2e
\usepackage{colm2024_conference}
\usepackage{graphicx}
\usepackage{subcaption}
\usepackage{tikz}
\usetikzlibrary{arrows.meta, positioning}
\usepackage[inkscapelatex=false]{svg}
\usepackage{authblk}

\usepackage{microtype}
\usepackage{hyperref}
\usepackage{amsmath}
\usepackage{url}
\usepackage{booktabs}
\usepackage{pifont}
\usepackage{xcolor}
\usepackage{multirow}
\definecolor{darkblue}{rgb}{0, 0, 0.5}
\hypersetup{colorlinks=true, citecolor=darkblue, linkcolor=darkblue, urlcolor=darkblue}

\title{Generating Leakage-Free Benchmarks for\\ Robust RAG Evaluation}

\author[1]{Jiayi Liu*}
\author[2]{Jiaxing Zhang}
\author[3]{Bowen Jin}
\author[1,4]{Jennifer Neville*}

\affil[1]{Department of Computer Science, Purdue University}
\affil[2]{New Jersey Institute of Technology}
\affil[3]{University of Illinois at Urbana-Champaign}
\affil[4]{Microsoft Research}

\affil[ ]{\texttt{\{liu2861, neville\}@purdue.edu, jz48@njit.edu, bowenj4@illinois.edu}}

\colmfinalcopy % Uncomment for camera-ready version, but NOT for submission.
\begin{document}

\maketitle

\begin{abstract}

Retrieval-augmented generation (RAG) is widely used to augment large language models (LLMs) with external knowledge. However, many benchmark datasets, designed to test RAG performance, comprise many questions that can already be answered from an LLM’s parametric memory. This leads to unreliable evaluation. We refer to this phenomenon as knowledge leakage---cases where RAG tasks are solvable without retrieval. This issue worsens over time due to benchmark aging. As benchmarks are reused for training, their contents are increasingly absorbed into model parameters, making them less effective for evaluating retrieval.

We introduce \textbf{SeedRG}, a semi-synthetic benchmark generation pipeline that mitigates knowledge leakage and addresses the issue of benchmark aging. Starting from a seed benchmark dataset, SeedRG extracts a reasoning graph from question–context pairs to capture their underlying reasoning structure, and then generates new examples via type-constrained entity replacement. This process produces structurally similar but novel instances that are unlikely to exist in the model’s parametric knowledge, while preserving the original reasoning patterns. To ensure quality, we incorporate two verification steps: (1) a reasoning-graph consistency check to maintain task difficulty, and (2) a knowledge-leakage filter to exclude instances answerable without retrieval.

We evaluate SeedRG on three seed benchmarks (HotpotQA, 2WikiMultihopQA, QASC) and three popular LLMs (GPT-5, Claude Sonnet 4.5, Gemini 2.5 Flash). SeedRG reduces knowledge leakage by at least 78\% while preserving reasoning difficulty. By removing the confounding effect of parametric knowledge, SeedRG reveals meaningful variability across RAG systems that is otherwise obscured. Prior benchmarks show uniformly high performance across RAG methods (HippoRAG, GraphRAG, OGRAG, SemanticRAG), because performance is dominated by model knowledge. In contrast, SeedRG surfaces clear differences in retrieval and reasoning ability across the methods. Beyond benchmark construction, we provide a systematic analysis linking reasoning difficulty to graph structure, showing how structural variations induce predictable changes in model accuracy. Together, these results demonstrate that SeedRG enables more discriminative and robust evaluation of RAG systems.

\end{abstract}

\section{Introduction}

%Retrieval-Augmented Generation (RAG) has become the one the most popular methods to improve LLM model accuracy by retrieving data from external sources. RAG utilizes the reasoning ability of current LLMs and doesn't need to retrain LLM itself with new knowledge. Multi-hop question answering benchmarks are the standard way to measure progress. But what if these benchmarks are not actually testing the efficacy of RAG? We find that on widely used multi-hop QA datasets, large language models answer 52--80\% of questions correctly \emph{without retrieving a single document}. When the retrieval component is no longer necessary for the majority of test questions, the benchmark cannot meaningfully distinguish retrieval systems: any method, no matter how poor its retrieval, inherits a high baseline from the model's parametric knowledge alone.

Retrieval-Augmented Generation (RAG)~\cite{lewis2020retrieval} is widely used to improve large language models (LLMs) by incorporating external knowledge at inference time. Because RAG avoids retraining by leveraging retrieval, it has become a standard approach for knowledge-intensive tasks. Progress on RAG systems is typically measured using multi-hop question answering benchmarks, which are intended to require retrieval over multiple documents.

However, these benchmarks often fail to test retrieval at all. We find that on widely used multi-hop QA datasets \cite{}, LLMs can answer even more than half of questions correctly \emph{without retrieving any documents}. When the majority of questions are solvable from LLM parametric memory alone, evaluation results are dominated by the model’s internal knowledge rather than the quality of retrieval. As a result, current benchmarks cannot meaningfully distinguish RAG systems---even weak retrievers inherit strong performance from the underlying model.

%This finding calls into question a large body of empirical work. RAG systems are now central to knowledge-intensive applications, from open-domain question answering~\cite{petroni2021kilt} to fact verification and domain-specific reasoning. A growing ecosystem of retrieval architectures---dense retrieval, graph-based retrieval, ontology-guided retrieval---competes on these benchmarks. Yet if the benchmarks themselves do not require retrieval, published rankings may reflect differences in the LLM's memorization rather than in retrieval quality.

This issue undermines a large body of empirical work. RAG systems are now central to applications such as open-domain question answering~\cite{petroni2021kilt}, fact verification, and domain-specific reasoning, with a growing ecosystem of retrieval methods---including dense, graph-based, and ontology-guided approaches---competing on shared benchmarks. When benchmarks can be solved from parametric memory, performance is dominated by the underlying model rather than the retriever, collapsing the observable differences between systems. This masks variation in retrieval quality and limits our ability to measure how different retrieval strategies influence downstream reasoning.

%We identify two root causes behind this evaluation failure:

%\begin{itemize}
%    \item \textbf{Knowledge Leakage.} The questions in existing benchmarks test widely known facts that large language models have already internalized during pretraining. The retrieval step adds no information that the model does not already possess, making the retriever's contribution unmeasurable.

%    \item \textbf{Benchmark Aging.} As pretraining corpora expand, static benchmarks are progressively absorbed into model weights. A dataset that was challenging at release becomes trivial for later-generation models---not because retrieval improved, but because the model memorized the answers.
%\end{itemize}

We identify two mechanisms that drive this evaluation failure. \textbf{Knowledge leakage} \cite{agarwal2024prompt,wu2025know,desai2026safegpt,yoon2025hypothetical} occurs when benchmark questions are answerable from an LLM’s parametric memory, making retrieval unnecessary. \textbf{Benchmark aging} \cite{zhou2023don,zhang2026benchmark} compounds this effect over time. As benchmarks are reused in training and data curation pipelines, their contents are absorbed into model parameters, progressively reducing their ability to test retrieval. Together, these mechanisms collapse the evaluation signal, obscuring differences between systems.

Addressing this problem requires benchmarks that are (1) outside the model’s parametric knowledge and (2) regenerable to remain robust to benchmark aging. A natural approach is to generate new data using LLMs. However, naive generation fails to meet these requirements: generated questions often reuse well-known entities (leading to continued leakage), may introduce factual errors in supporting context, and provide no control over reasoning difficulty.

% Addressing this problem requires benchmarks that are (1) outside the model’s parametric knowledge and (2) regenerable to remain robust to benchmark aging. A natural approach is to generate new data using LLMs. However, naive {\em direct generation} fails to meet these requirements. Generated questions tend to reuse well-known entities, resulting in continued leakage---25.3\% of generated questions are still answerable without any context. In addition, 13.3\% of generated supporting passages are plausible-sounding but factually incorrect, undermining the integrity of the benchmark. More broadly, unconstrained generation provides no control over reasoning difficulty, leading to instances that are either trivially easy or inconsistently hard.

%Together, these two mechanisms create a cycle: the reliability of benchmarks will degrade over time, and new RAG methods are evaluated against increasingly uninformative baselines. Breaking this cycle requires benchmarks that are \emph{provably outside} the LLM's parametric knowledge and that can be \emph{regenerated on demand} to stay ahead of benchmark aging. 

% \begin{figure}[t]
%     \centering
%     \includegraphics[width=\linewidth]{figure/problem.svg}
%     \caption{The RAG evaluation gap. \textbf{(a)} Existing benchmarks overlap with LLM pretraining data, making retrieval redundant. \textbf{(b)} SeedRG preserves the reasoning structure of seed questions while replacing all entities with novel counterparts, ensuring no overlap with parametric knowledge.}
%     \label{fig:rag_dataset}
% \end{figure}

\begin{figure}[t!]
    \centering
    \includegraphics[width=0.9\columnwidth]{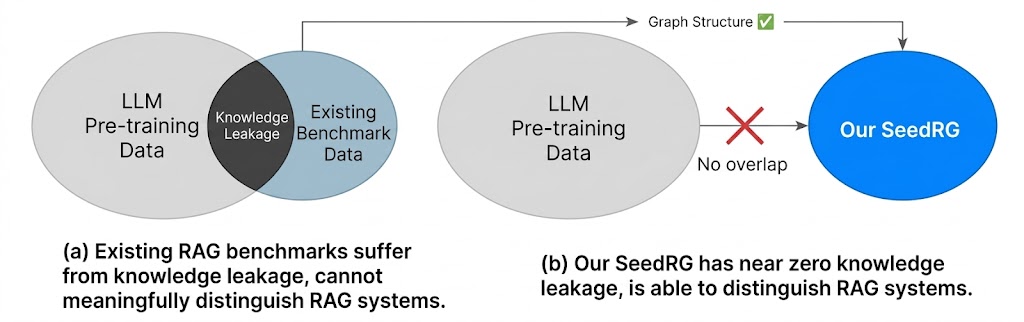}
    \caption{An example of the RAG evaluation gap. \textbf{(a)} Existing benchmarks overlap with LLM pretraining data, making retrieval redundant. \textbf{(b)} SeedRG preserves the reasoning structure of seed questions while replacing all entities with novel counterparts, ensuring no overlap with parametric knowledge.}
    \label{fig:rag_dataset}
\end{figure}
%A natural idea is to ask the LLM to generate entirely new questions and contexts from scratch. However, unconstrained generation introduces two subtle failure modes. First, the LLM tends to produce questions using the entities it knows well, we find that 25.3\% of directly generated questions are still answerable without any context. Second, the LLM may generate plausible-sounding but factually incorrect supporting passages at a rate of 13.3\%, undermining the integrity of the benchmark itself. Furthermore, free-form generation provides no control over reasoning difficulty: without explicit structural constraints, the generated questions may be trivially easy or impossibly hard, making cross-system comparisons unreliable. What should be generated is: question-context pairs that could not be answered by any current knowledge, and does not stand against any real-world knowledge, while answering it does not change the difficulty.

%We propose \textbf{SeedReasoningGraph}(SeedRG), a benchmark generation pipeline that starts from an existing multi-hop benchmark and transform each question-context pair through two controlled operations: (1) \textit{reasoning graph extraction}, which using question-context graph structure as a fingerprint to preserve the orginal reasoning difficulty; and (2) \textit{type-constrained entity replacement}, which substitutes every entity with a novel counterpart of the same semantic type (e.g., a composer is replaced only by another composer), to produce questions that are provably outside the LLM's training distribution. 

We introduce \textbf{SeedRG}, a semi-synthetic benchmark generation pipeline that addresses these challenges. Starting from a seed benchmark, SeedRG extracts a \emph{reasoning graph} from each question–context pair to capture its underlying structure, and then generates new examples via \emph{type-constrained entity replacement}. This produces structurally equivalent but novel instances that are unlikely to exist in the model’s parametric knowledge, while preserving the original reasoning patterns.

To ensure quality, SeedRG incorporates two verification steps. A \emph{reasoning graph consistency check} ensures that the transformed examples preserve the original reasoning structure and difficulty. A \emph{knowledge leakage check} filters out instances that can be answered without retrieval. Together, these steps ensure that generated examples are both retrieval-dependent and difficulty-preserving. We further introduce two metrics---\emph{leakage error} and \emph{answerability accuracy}---to quantify the effectiveness of RAG benchmarks.

%We also assign two verification steps to close the loop. We apply \emph{reasoning graph check} to confirm the matchness between new question-context pair and old ones, which prevents uncontrolled difficulty shifts. Another verification is \emph{knowledge leakage check}, which confirms that the LLM cannot answer the transformed question without the supporting context. Together, these safeguards ensure that every accepted new sample is retrieval-dependent and difficulty-preserving. To measure the effectiveness of the RAG benchmark, we introduce two criteria that any valid RAG benchmark should optimize towards---\emph{leakage error} and \emph{answerability accuracy}.

In summary, our contributions are as follows:

\begin{enumerate}
    \item We provide systematic evidence that three widely used multi-hop QA benchmarks suffer from \textbf{knowledge leakage}, and formalize benchmark quality in terms of leakage error and answerability accuracy.

    \item We propose \textbf{SeedRG}, a semi-synthetic pipeline that generates leakage-free benchmarks by combining reasoning graph extraction with type-constrained entity replacement, along with dual verification to preserve difficulty and enforce retrieval dependence.

    \item We show that SeedRG produces benchmarks that reduce knowledge leakage, preserve reasoning difficulty, and reveal meaningful performance differences across RAG systems that are obscured in existing benchmarks. Compared to direct LLM generation, SeedRG yields higher-quality benchmarks with substantially lower leakage and fewer factual inconsistencies.
\end{enumerate}

\section{Background}
\subsection{Synthetic datasets}
\paragraph{Synthetic Benchmarks for RAG}Recent frameworks have standardized the evaluation of RAG systems through automated data synthesis. RAGEval~\cite{zhu2024rageval} generates schema-driven datasets to assess \emph{scenario-specific} factual accuracy. It defines dataset quality using three key metrics: \emph{completeness} of the answer, absence of \emph{hallucination}, and \emph{irrelevance} of non-essential content. To scale this approach, BenchmarkQED~\cite{benchmarkqed2025} employs the AutoQ tool to generate synthetic queries across a principled $2\times2$ taxonomy. It measures quality via \emph{coverage} (diversity of query types) and \emph{rigor} (stability of comparative rankings). However, both frameworks share a critical limitation: they fail to add restrictions that avoid generating queries already present in the LLM's knowledge. By not explicitly disentangling parametric memory from retrieval necessity, these benchmarks struggle to isolate the true utility of the retrieval component.\paragraph{Synthetic Data for Instruction Tuning}In the broader context of model alignment, synthetic data quality is often defined by downstream efficiency rather than retrieval isolation. Distilling Step-by-Step~\cite{hsieh2023distilling} and Orca~\cite{mukherjee2023orca} demonstrate that "good" synthetic data allows smaller student models to achieve \emph{teacher-parity} with significantly fewer training samples. Similarly, MetaMath~\cite{yu2023metamath} and SynPO~\cite{dong2024self} validate dataset quality through \emph{reasoning transfer} to unseen math tasks and \emph{iterative win-rate improvements} on public leaderboards. While these methods successfully enhance general reasoning and alignment, they do not address the specific knowledge-boundary constraints required to rigorously prevent memory-based hallucinations in RAG tasks.

We observe no significant improvement in accuracy across several benchmarks. In some cases, performance even degrades, despite the retrieval module returning correct supporting paragraphs. This suggests that the language model often already possesses sufficient parametric knowledge to answer the questions without relying on retrieved evidence, limiting the benefit of external retrieval.

\subsection{Knowledge Leakage in LLM}
Knowledge leakage has been widely discussed in previous research. Recent work has doubted the efficacy of LLM in recommendation~\cite{zhang2026benchmark,zhou2023don}, query expansion~\cite{yoon2025hypothetical}, privacy~\cite{agarwal2024prompt,wu2025know,desai2026safegpt}, and many other tasks~\cite{baser2025thinkeval}. 

To prevent the knowledge leakage, researchers applied different stratigies. ~\cite{agarwal2024prompt, desai2026safegpt} use defense instructions and filtering guardrails within the prompt or system layer to block leakage.
To prevent the knowledge leakage, researchers applied different stratigies. ~\cite{agarwal2024prompt, desai2026safegpt} mitigate the problem via prompt to block leakage. ~\cite{wu2025know} and ~\cite{baser2025thinkeval} propose secure KV-cache management and knowledge graph monitoring as extra system design to track and solve leakage issues. ~\cite{zhou2023don, zhang2026benchmark} added constraints in fine-tuning stage to prevent knowledge leakage.

%\section{What Makes a Valid RAG Benchmark?}
\section{Formalizing Valid RAG Benchmarks}
\label{sec:formal_obj}

%Before proposing a solution, we formalize what it means for a RAG benchmark to be \emph{valid}: to actually measure retrieval quality rather than pre-existing knowledge recall. We first define two necessary conditions, then show what existing benchmarks lack.

%Evaluating the efficacy of RAG algorithms requires a clear distinction between what a LLM model knows and what the RAG algorithm retrieves. Current benchmarks suffer from a 'knowledge leakage' problem: LLM could be able to answer the question in RAG benchmarks even without any retrieval. This phenomenon makes it nearly impossible to attribute performance gains to the retrieval mechanism itself. We identify two major drivers of our motivation: \textbf{Knowledge Leakage} and \textbf{Benchmark Aging}.

Before proposing a solution, we formalize what it means for a RAG benchmark to be \emph{valid}. Specifically, it should measure retrieval quality rather than parametric knowledge recall. We identify two necessary conditions for validity and show how existing benchmarks violate them.

A valid RAG benchmark must separate what the model already knows from what is provided through retrieval. When this separation fails, evaluation signal collapses and performance is dominated by parametric knowledge, making it difficult to attribute gains to retrieval. We formalize this failure as \textbf{knowledge leakage}, and argue that it is exacerbated over time due to \textbf{benchmark aging}.

% Existing benchmarks violate these criteria through two mechanisms (Figure~\ref{fig:redundancyinlfation}):

% \begin{figure}[t]
%     \centering
%     \includegraphics[width=0.8\linewidth]{figure/redundancy+inflation.png}
%     \caption{Two failure modes of existing RAG benchmarks. \textbf{Knowledge Leakage:} the LLM's parametric knowledge already contains the answer, making retrieval unnecessary. \textbf{Benchmark Aging:} static test sets are absorbed into expanding pretraining corpora, progressively eroding leakage error.}
%     \label{fig:redundancyinlfation}
% \end{figure}

\subsection{Knowledge Leakage}
%Let $M$ denote a language model and $\mathcal{Q}$ a benchmark dataset. For each question $q \in \mathcal{Q}$, we measure accuracy under two conditions: $\text{Acc}_{\text{no\_ctx}}$ (LLM answers with no context), $\text{Acc}_{\text{gold}}$ (LLM answers given the ground-truth document).

Let $M$ denote a language model and $\mathcal{Q}$ a benchmark dataset. For each question $q \in \mathcal{Q}$, we measure accuracy under two conditions: $\text{Acc}_{\text{no\_ctx}}$ (when the model answers are produced without any retrieved context) and $\text{Acc}_{\text{gold}}$ (when the model answers are produced given only the ground-truth supporting documents).

%When a RAG algorithm test with benchmark that consist of commonly known facts, the LLM can answer from pre-existing knowledge alone, since the LLM has already been trained with those RAG algorithms. As we show in Section~\ref{sec:new_vs_old}, $\text{Acc}_{\text{no\_ctx}}$ reaches 52\% on HotpotQA~\cite{yang2018hotpotqa}, 56\% on 2WikiMultihopQA~\cite{welbl2018constructing}, and 80\% on QASC~\cite{khot2020qasc}. Based on that, we would never know whether the accuracy of RAG algorithms are brought by the efficacy of retrieval or the things that LLM already learn.

Knowledge leakage occurs when $\text{Acc}_{\text{no\_ctx}}$ is high, indicating that questions can be answered directly from parametric memory. In this regime, retrieval has provided no additional information, and benchmark performance no longer reflects retrieval quality. As we show in Section~\ref{sec:new_vs_old}, $\text{Acc}_{\text{no\_ctx}}$ reaches 52\% on HotpotQA~\cite{yang2018hotpotqa}, 62\% on 2WikiMultihopQA~\cite{welbl2018constructing}, and 75\% on QASC~\cite{khot2020qasc}.

%\begin{enumerate}
    %\item \textbf{Leakage Error.} We need to ensure the questions cannot be answered by the LLM alone, reducing the chance of prior knowledge as much as possible.
    %\[
    %    \min \text{Acc}_{\text{no\_ctx}}(\mathcal{Q})
    %\]

    %\item \textbf{Answerability Accuracy.} Answer accuracy should improve substantially when the correct gold context is provided, versus when no context is given.
    %\[
    %    \max (\text{Acc}_{\text{gold}}(\mathcal{Q}) - \text{Acc}_{\text{no\_ctx}}(\mathcal{Q}))
    %\]
%\end{enumerate}

We formalize two criteria for a valid RAG benchmark:

\begin{enumerate}
    \item \textbf{Leakage Error.} The extent to which questions are answerable from parametric knowledge alone:
    \[
        \text{Acc}_{\text{no\_ctx}}(\mathcal{Q})
    \]
    A valid benchmark should have low leakage error.

    \item \textbf{Answerability Accuracy.} The improvement in accuracy when the correct context is provided:
    \[
        \text{Acc}_{\text{gold}}(\mathcal{Q}) - \text{Acc}_{\text{no\_ctx}}(\mathcal{Q})
    \]
    A valid benchmark should exhibit high answerability accuracy.
\end{enumerate}

%\noindent We aim to minimize leakag error and maximize answerability accuracy, thereby ensuring the RAG datasets could validate the performance of RAG algorithms.
\noindent Together, these criteria ensure that benchmark performance reflects retrieval-dependent reasoning rather than memorization.

\subsection{Benchmark Aging} 
%Even if a benchmark has not been put into the knowledge base of LLM yet, it degrades over time~\cite{zhou2023don,zhang2026benchmark}: each new generation of LLMs trains on a larger slice of the web, progressively absorbing test questions into their storage. For any frozen benchmarks, there is no mechanism to combat the erosion. The result is a silent inflation of reported scores that reflects better LLM memorization, not retrieval improvement.

Even when a benchmark initially satisfies these criteria, it degrades over time~\cite{zhou2023don,zhang2026benchmark}. As pretraining corpora expand, benchmark questions are increasingly incorporated into model training data. This progressively increases $\text{Acc}_{\text{no\_ctx}}$, increasing leakage error and weakening the benchmark’s ability to test retrieval.

For static benchmarks, this process is irreversible---there is no mechanism to prevent their absorption into parametric knowledge. The result is a gradual inflation of reported performance that reflects improved memorization rather than improved retrieval. As a consequence, benchmarks lose their discriminative power to assess RAG systems. A valid RAG benchmark must therefore not only minimize knowledge leakage, but also remain robust to benchmark aging.

%% ================================================================
\section{Methodology}
\label{sec:pipeline}

%Rather than introducing a single static benchmark, we propose \textbf{SeedRG}, a framework for \textbf{continually generating} benchmarks which enforce \textbf{leakage error and answerability accuracy}, thereby mitigating \textbf{benchmark aging}.
Rather than introducing a static benchmark, we propose \textbf{SeedRG}, a framework for \textbf{continually generating} benchmarks that minimize \textbf{knowledge leakage} while preserving \textbf{answerability accuracy}, thereby mitigating \textbf{benchmark aging}.

%SeedRG uses a multi-hop RAG benchmark as a seed to produce new contexts and questions. To tackle with the challenges we stated above, SeedRG preserves the reasoning graph of the original questions-contexts, but replaces the entities with ones that are completely outside the LLM’s internal knowledge. We propose the following tasks:

SeedRG takes a multi-hop RAG benchmark as a seed and transforms each question–context pair into a new instance that is retrieval-dependent and difficulty-preserving. The key idea is to preserve the underlying reasoning structure while replacing entities with ones that fall outside the model’s parametric knowledge.

We formalize three requirements for valid benchmark generation:

\begin{enumerate}
    %\item \textbf{Leakage eror and answerability accuracy.} Generated questions must fall outside the LLM's parametric knowledge (leakage error), while remaining answerable with the correct context (answerability accuracy). This requires both a generation mechanism that produces novel content and a verification step that rejects any question within the LLM knowledge base.
    \item \textbf{Leakage and Answerability.} Generated questions must not be answerable from parametric knowledge alone, while remaining answerable given the correct context. This requires both a generation mechanism that produces novel content and a verification step that filters out leaking instances.

    %\item \textbf{Renewable generation.} The framework must produce fresh benchmark instances on demand, so that even if a particular set of questions is eventually absorbed into future training data, a new set can replace it immediately. This makes the benchmark inherently resistant to aging.
    \item \textbf{Renewable Generation.} The framework must produce fresh benchmark instances on demand, so that as models absorb existing data, new instances can replace them. This ensures robustness to benchmark aging.

    %\item \textbf{Reasoning difficulty preservation.} The generation process must preserve reasoning difficulty: without explicit structural constraints, generated questions risk being trivially easy or impossibly hard, rendering cross-system comparisons meaningless. We address this through reasoning graph extraction (Section~\ref{sec:graph_extraction}).
    \item \textbf{Difficulty Preservation.} The generation process must preserve reasoning difficulty. Without structural constraints, generated questions may become trivially easy or arbitrarily hard, confounding retrieval performance with the model’s reasoning ability.
\end{enumerate}

%The overall framework is shown in Figure~\ref{fig:pipeline_structure}, which consists of two main components: type-constrained entity replacement (Section~\ref{sec:entity_replacement}) to \textbf{renewable} generate \textbf{retrieval-dependent} samples, and reasoning graph extraction (Section~\ref{sec:graph_extraction}) to \textbf{preserve the reasoning difficulty}.

\begin{figure}[t!]
    \centering
    \includegraphics[width=0.9\columnwidth]{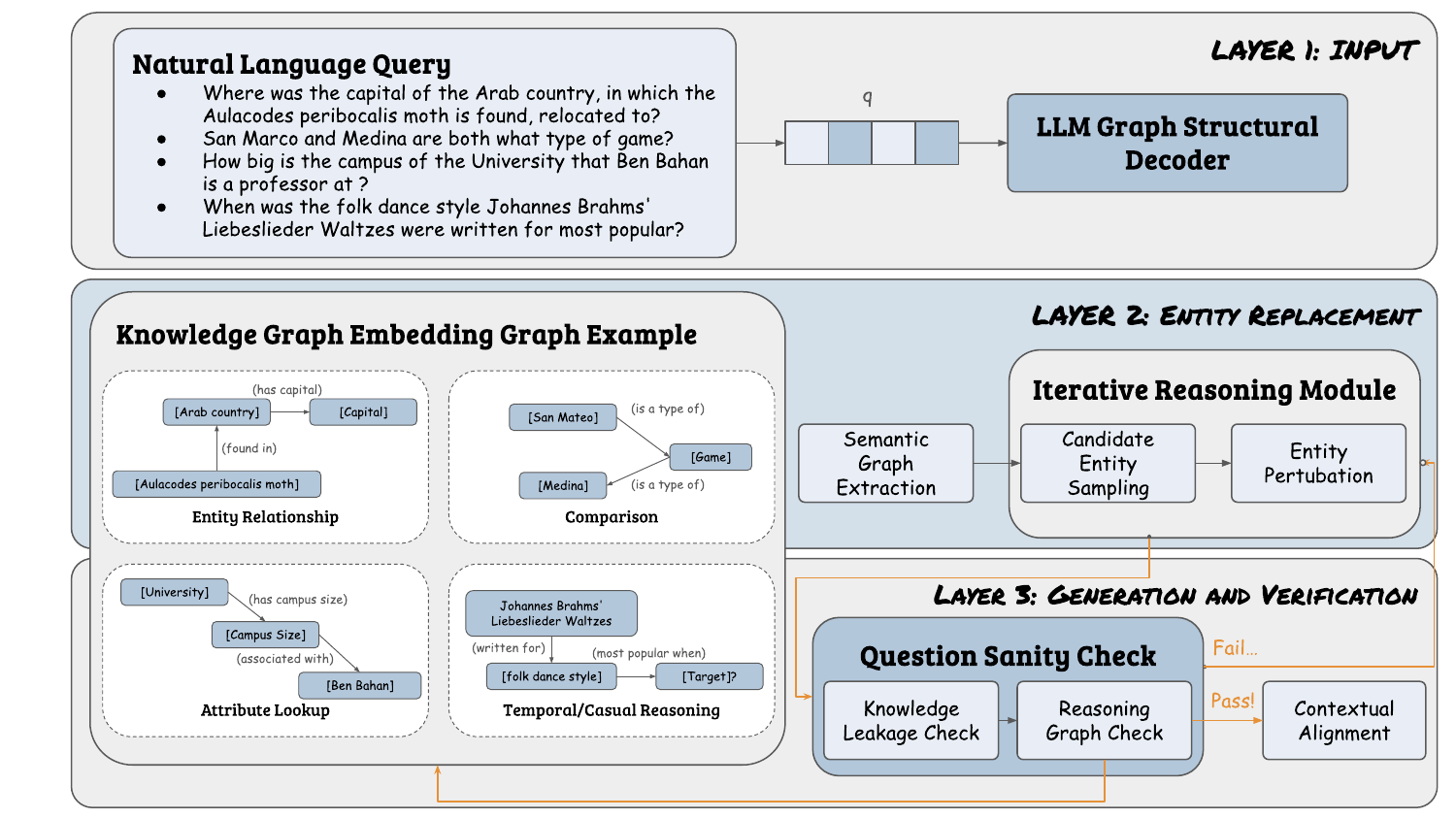}
    \caption{
    %Overview of our counterfactual dataset construction framework. Given a seed question, the pipeline extracts its reasoning graph, applies type-constrained entity replacement, and validates the result through reasoning graph and knowledge leakage checks. Samples that fail either check are rejected and regenerated.
    Overview of the SeedRG benchmark generation pipeline. Given a seed question–context pair, the pipeline extracts its reasoning graph, applies type-constrained entity replacement, and validates the result through reasoning graph and knowledge leakage checks. Samples that fail either check are rejected and regenerated.}
    \label{fig:pipeline_structure}
\end{figure}

The overall framework is shown in Figure~\ref{fig:pipeline_structure}. SeedRG consists of two main components: \textbf{type-constrained entity replacement} (Section~\ref{sec:entity_replacement}), which generates \textbf{retrieval-dependent} samples, and \textbf{reasoning graph extraction} (Section~\ref{sec:graph_extraction}), which ensures \textbf{difficulty preservation}.

\subsection{Type-Constrained Entity Replacement}
\label{sec:entity_replacement}

SeedRG replaces entities in both the question and context while ensuring the resulting instance is non-leaking and answerable.

% \subsubsection{Entity Sampling}
% \label{sec:entity_sampling}

%Given a seed question $q$ with context $c$ and answer $a$, we first extract the reasoning graph to identify all entity nodes. For each entity $e_i$, we prompt an LLM with: \textit{``Replace the name [entity] with another random name. The original and new name should be the same type, but the new one should not be famous---it could even be created by you.''} This produces a type-constrained substitute $e'_i$ (e.g., a composer replaced by another composer, a city by another city) that is obscure enough to fall outside the LLM's parametric knowledge.
Given a seed question $q$ with context $c$ and answer $a$, we first extract its reasoning graph to identify entity nodes. For each entity $e_i$, we prompt an LLM to generate a replacement of the same semantic type but unlikely to be present in parametric knowledge (e.g., a composer replaced by another composer, a city by another city). This produces a type-constrained substitute $e'_i$.

% \subsubsection{Joint Question-Context Transformation}
% \label{sec:joint_transform}

%The entity mapping $\mathcal{M}: \{e_i\} \to \{e'_i\}$ is applied simultaneously to the question, answer, and context: $q' = \mathcal{M}(q)$, $a' = \mathcal{M}(a)$, $c' = \mathcal{M}(c)$. For the context, we perform entity replacement at two levels. First, we directly substitute entity names in the original text, preserving the original prose and information density. Second, we extract knowledge graph triplets $\mathcal{T} = \{(s_i, r_i, o_i)\}$ from $c$, apply $\mathcal{M}$ to all entity mentions in the triplets, and regenerate a natural-language passage from the mapped triplets. This triplet-mediated path is used for the graph structure perturbation experiments (Section~\ref{sec:graph_structure_exp}), where we need to modify the graph topology while keeping the same entities.

We define a mapping $\mathcal{M}: \{e_i\} \to \{e'_i\}$ and apply it jointly to the question, answer, and context: $q' = \mathcal{M}(q)$, $a' = \mathcal{M}(a)$, and $c' = \mathcal{M}(c)$. 

For the context, we perform entity replacement in two ways. First, we directly substitute entity mentions in the original text, preserving surface form and information density. Second, we extract knowledge graph triplets $\mathcal{T} = \{(s_i, r_i, o_i)\}$ from $c$, apply $\mathcal{M}$ to all entities, and regenerate a natural-language passage from the transformed triplets. This second path enables controlled perturbations of graph structure (Section~\ref{sec:graph_structure_exp}).

% \subsubsection{Knowledge Leakage Check}
% \label{sec:leakage_check}

%A generated sample is only useful if the question cannot be answered without context. For each candidate $(q', a')$, we query the LLM with $q'$ alone (no context) $N=3$ times. If any response contains the correct answer $a'$, the sample is rejected and the pipeline re-samples with an additional constraint excluding previously tried names. This loop continues until a non-leaking replacement is found. Because each seed question can yield many valid entity mappings, SeedRG supports renewable generation: fresh benchmarks can be produced on demand, directly addressing benchmark aging.

A generated sample is only valid if the question cannot be answered without context. For each candidate $(q', a')$, we query the LLM with $q'$ alone (no context) multiple times ($N=3$). If any response contains the correct answer $a'$, the sample is rejected. The pipeline then resamples entity replacements with additional constraints to avoid previously tried entities. This process repeats until a non-leaking instance is obtained.

Because each seed question admits many valid replacements, this process supports renewable generation—new benchmark instances can be produced on demand, directly addressing benchmark aging.
\subsection{Reasoning Graph Extraction}
\label{sec:graph_extraction}
%The difficulty of a multi-hop question depends on how entities are connected: the number of hops, the path from question to answer, the branching structure. If entity replacement accidentally changes these connections, the question becomes easier or harder. To prevent this, we extract reasoning graphs before and after replacement and check that they match.
The difficulty of a multi-hop question is determined by its reasoning structure—the number of hops, the dependency chain, and the connectivity between entities. If the structure of the reasoning graph changes, it might result in different reasoning difficulties.

To preserve reasoning difficulty, we extract reasoning graphs before and after transformation. For a seed question $q$ with context $c$, we construct a question graph $G_q = (V_q, E_q)$ that captures the reasoning chain, and a context graph $G_c = (V_c, E_c)$ derived from factual triplets $\mathcal{T} = \{(s_i, r_i, o_i)\}$. After transformation, we extract $G_{q'}$ and $G_{c'}$ and verify structural equivalence: $G_{q'} \cong G_q$ and $G_{c'} \cong G_c$. If the structure is not preserved, the sample is discarded and regenerated.

%For a seed question $q$ with context $c$, we extract two graphs. The question graph $G_q = (V_q, E_q)$ captures the reasoning chain---for example, \texttt{[album] --(debut of)--> [band] --(formed by)--> [?]}---by prompting an LLM to output a structured node-edge representation. The context graph $G_c = (V_c, E_c)$ is extracted by decomposing $c$ into factual triplets $\mathcal{T} = \{(s_i, r_i, o_i)\}$, capturing how entities relate to each other in the supporting evidence.

%After entity replacement produces $q'$ and $c'$, we extract $G_{q'}$ and $G_{c'}$ using the same procedure and verify that the graph structure is preserved: $G_{q'} \cong G_q$ and $G_{c'} \cong G_c$. If not, the pipeline retries until the structure matches. We show in Section~\ref{sec:graph_structure_exp} that this check is necessary---changing graph structure directly changes reasoning difficulty.

We also prove our hypothesis in Section~\ref{sec:graph_structure_exp}, that modifying the structure directly changes task difficulty. Hence, the reasoning graph verification step is necessary to maintain the reasoning difficulty.

\section{Experiments}

\subsection{Experimental Setup}

%All generated benchmarks are generated using GPT-4o-mini using the algorithm we mentioned in Section \ref{sec:pipeline}. The performance of benchmarks is evaluated with three stronger reasoning LLMs: GPT-5, Claude Sonnet 4.5, and Gemini 2.5 Flash.\footnote{Our code is available at \url{https://anonymous.4open.science/r/SeedRG-02D6}.}

All generated benchmarks are produced using GPT-4o-mini following the SeedRG pipeline described in Section~\ref{sec:pipeline}. We evaluate benchmark quality using three frontier LLMs with strong reasoning capabilities: GPT-5, Claude Sonnet 4.5, and Gemini 2.5 Flash.\footnote{Our code is available at \url{https://anonymous.4open.science/r/SeedRG-02D6}.}

Our experiments address two questions. First, we evaluate whether SeedRG produces higher-quality benchmarks compared to Direct Generation (DG) and the original benchmarks, in terms of knowledge leakage and answerability accuracy. Second, we use these benchmarks to evaluate RAG systems and examine whether SeedRG reveals differences in retrieval performance that are obscured in existing benchmarks.

We compare SeedRG against \textbf{Direct Generation (DG)}, where an LLM generates new question–context–answer triples from scratch given a seed example. The DG prompt explicitly instructs the model to (1) preserve reasoning type and difficulty, (2) use entirely different entities and facts, and (3) require the provided context for answering (i.e., not answerable from parametric knowledge alone).

To evaluate both benchmark quality and retrieval dependence, we measure performance under the following conditions/methods:

%We compare SeedRG against Direct Generation (DG), where LLM generates entirely new question--context--answer triples from scratch given only the original question and context as a seed. Importantly, the DG prompt explicitly instructs the LLM to (1) \textit{``have the same reasoning type and difficulty level,''} (2) \textit{``be about completely different entities, topics, and facts,''} and (3) \textit{``require the context to answer (not answerable from general knowledge alone).''}.

%We test the efficacy of SeedRG with seven retrieval conditions:
\begin{itemize}
    \item \textbf{No-Context}: %The LLM answers without any retrieved context, the accuracy here indicates knowledge leakage and also the lower bound.
    The LLM answers the question without any retrieved context. This measures knowledge leakage.

    \item \textbf{Gold}: %The ground-truth supporting document is provided as the only context to indicate the upper bound on context-dependent accuracy.
    The ground-truth supporting documents are provided as context, representing an upper bound on answerability.

%     \item \textbf{Together}: All corpus documents from the sampled question pool are concatenated and provided as context.
%     TBA: change together to all context, call old orig

% TBA: Change SeedRG to SeedRG, update it in our intro
% Add more explanation for hallucination. We have 0, because perturbation makes it so far from its previous distribution model knowledge, while DG still generates inside the model knowledge.

% TBA: change the figure 3c, complexity of the new questions -contexts. Flip up to be complexity change. If perturb the edges, it will make the question harder for model to answer even give the golden context. Don't call it accuracy. 

% One generate questions-contexts with same complexity, the other one generate questions-contexts with different complexity. The way define complexity is diff in gold accuracy.

% We use red to test reasoning. SeedRG no structrue change, seedRG with structure change. For future work red could be reasoning construction.

    \item \textbf{HippoRAG}~\cite{gutierrez2024hipporag} %is a RAG framework that mimicks human memory with an long-term memory for LLMs. It pre-processes inputs by transforming the corpus into a knowledge graph as hippocampal index.
    is a RAG framework that mimics human memory by constructing a knowledge graph as a long-term memory index.

    \item \textbf{OGRAG}~\cite{sharma2025og} %is an Ontology-Grounded RAG algorithm. OGRAG first ask domain experts to get the ontology, then build documents with a ontology-based hypergraph.
    is an ontology-grounded RAG approach that organizes documents using expert-defined ontologies and hypergraph structures.

    \item \textbf{GraphRAG}~\cite{edge2024local} %is a graph-based RAG approach. By taking the whole knowledge as a large text corpus, GraphRAG uses LLM to construct knowledge graph of the whole corpus.
    constructs a global knowledge graph over the corpus and performs retrieval over graph-structured representations.

    \item \textbf{SemanticRAG} %is a simple RAG algorithm that uses OpenAI embeddings for corpus and question, than rank with cosine similarity ranking. It represents the most common production RAG approach.
    is a standard embedding-based retrieval approach that ranks documents using cosine similarity between query and document embeddings, representing a common production baseline.

\end{itemize}

% ============================================================
% SECTION 1: Ours vs Original vs DG (Pure, Together, Gold-Pure)
% ============================================================
\subsection{Benchmark Comparison}
\label{sec:new_vs_old}
%We compare three benchmark: our SeedRG, baseline Direct Generation (DG), and the Original unmodified benchmark across three key metrics in Table~\ref{tab:criteria}. We verified Leakage Error criteria via $\text{Acc}_{\text{no\_ctx}}(\mathcal{Q})$ (lower is better), and answerability accuracy via $\text{Acc}_{\text{gold}}(\mathcal{Q}) - \text{Acc}_{\text{no\_ctx}}(\mathcal{Q})$ (higher is better).

We compare three benchmarks: SeedRG, Direct Generation (DG), and the original benchmark across two metrics defined in Section~\ref{sec:formal_obj}: leakage error, measured by $\text{Acc}_{\text{no\_ctx}}(\mathcal{Q})$ and answerability accuracy, measured by $\text{Acc}_{\text{gold}}(\mathcal{Q}) - \text{Acc}_{\text{no\_ctx}}(\mathcal{Q})$ (higher is better). Results are summarized in Table~\ref{tab:criteria}.

\begin{table*}[t]
\centering
\footnotesize
\setlength{\tabcolsep}{4pt}
\caption{
%Benchmark comparison. \textbf{Leakage Error} ($\text{Acc}_{\text{no\_ctx}}$, lower is better): questions should not be answerable without context. \textbf{Answerability Accuracy} ($\text{Acc}_{\text{gold}} - \text{Acc}_{\text{no\_ctx}}$, higher is better): the gap quantifies how much retrieval contributes.
Benchmark comparison. \textbf{Leakage Error} ($\text{Acc}_{\text{no\_ctx}}$, lower is better): fraction of questions answerable without context. \textbf{Answerability Accuracy} ($\text{Acc}_{\text{gold}} - \text{Acc}_{\text{no\_ctx}}$, higher is better): potential gain from retrieval.
}
\label{tab:criteria}
\begin{tabular}{ll ccc ccc ccc}
\toprule
& & \multicolumn{3}{c}{\textbf{GPT-5}} & \multicolumn{3}{c}{\textbf{Claude Sonnet 4.5}} & \multicolumn{3}{c}{\textbf{Gemini 2.5 Flash}} \\
\cmidrule(lr){3-5} \cmidrule(lr){6-8} \cmidrule(lr){9-11}
\textbf{Metric} & \textbf{Dataset} & SeedRG & DG & Orig & SeedRG & DG & Orig & SeedRG & DG & Orig \\
\midrule
\multirow{3}{*}{\shortstack[l]{Leakage Error}}
 & HotpotQA & \textbf{.014} & .190 & .500 & \textbf{.135} & .310 & .459 & \textbf{.041} & .180 & .311 \\
 & WikiHop  & \textbf{.114} & .390 & .629 & \textbf{.129} & .420 & .386 & \textbf{.057} & .350 & .400 \\
 & QASC     & \textbf{.140} & .320 & .750 & \textbf{.170} & .530 & .820 & \textbf{.160} & .380 & .770 \\
\midrule
\multirow{3}{*}{\shortstack[l]{Answerability Accuracy}}
 & HotpotQA & .418 & \textbf{.500} & .338 & \textbf{.595} & .470 & .473 & .473 & .460 & \textbf{.500} \\
 & WikiHop  & \textbf{.457} & .330 & .200 & \textbf{.657} & .460 & .543 & \textbf{.514} & .380 & .414 \\
 & QASC     & \textbf{.710} & .410 & .180 & \textbf{.690} & .290 & .170 & \textbf{.760} & .380 & .210 \\
\bottomrule
\end{tabular}
\end{table*}

\paragraph{Leakage Error.}
%On the original benchmarks, all three QA engines achieve high no-context accuracy (31--78\%), confirming severe knowledge leakage. SeedRG reduces no-context accuracy dramatically---GPT-5 to 1.4\% on HotpotQA, Gemini to 4.1\%, Claude to 13.5\%---satisfying leakage error across all engines. Direct Generation also reduces no-context accuracy relative to Old, but to a lesser degree (18--53\%), indicating partial but insufficient knowledge leakage elimination.
On the original benchmarks, all three LLMs achieve high no-context accuracy (31--78\%), confirming substantial knowledge leakage. In contrast, SeedRG reduces no-context accuracy dramatically—to 1.4\% (GPT-5) on HotpotQA, 4.1\% (Gemini), and 13.5\% (Claude)—effectively eliminating leakage across models. Direct Generation also reduces leakage relative to the original benchmarks, but only partially (18--53\%), indicating that naive generation is insufficient to remove parametric shortcuts.

\paragraph{Answerability Accuracy.}
%The $\text{Acc}_{\text{gold}} - \text{Acc}_{\text{no\_ctx}}$ gap directly quantifies how much retrieval contributes. SeedRG consistently produces the largest gaps (42--76\%), indicating that SeedRG benchmarks maximally differentiate retrieval quality. Old benchmarks show the smallest gaps (18--54\%) due to high no-context accuracy, while Direct Generation falls in between.

The gap $\text{Acc}_{\text{gold}} - \text{Acc}_{\text{no\_ctx}}$ quantifies the potential contribution of retrieval, ie. how much accuracy could improve if retrieval were perfect. SeedRG consistently yields the largest gaps (42--76\%), indicating that these benchmarks create substantial headroom for retrieval to matter. In contrast, the original benchmarks exhibit smaller gaps (18--54\%) due to high no-context accuracy, while Direct Generation again lies in between.
Taken together, these results show that SeedRG restores evaluation signal by reducing leakage and increasing the extent to which performance can depend on retrieval.

% ============================================================
% SECTION 2: Quality Analysis (Knowledge Leakage + Hallucination)
% ============================================================
\subsection{Generation Quality Analysis}
\label{sec:quality_analysis}
\subsubsection{Generation Quality for Question and Context}

%In this section, we compare the quality for question and context through three dimensions: The knowledge leakage, factual hallucination, and resoning difficulty. The resulst are avilable in Figure~\ref{fig:quality_perturbation}.
We evaluate generation quality along three dimensions: knowledge leakage, factual consistency, and reasoning difficulty. Results are shown in Figure~\ref{fig:quality_perturbation}.

%\paragraph{Knowledge leakage and hallucination.} We query the LLM with each generated question and no context. If it answers correctly, the question leaks. If it gives a confident but wrong answer that contradicts the generated context, the context is hallucinated. Otherwise the question is non-leaking.
\paragraph{Knowledge leakage and factual consistency.} For each generated question, we query the LLM without context. If the model answers correctly, the question exhibits knowledge leakage. If the model produces a confident but incorrect answer that contradicts the generated context, the context is deemed factually inconsistent. Otherwise, the question is considered non-leaking.
Figure~\ref{fig:quality_perturbation}(a) shows that Direct Generation (DG) produces low-quality benchmarks: only 0--11\% of questions are non-leaking, while 19--39\% leak and 53--70\% exhibit factual inconsistencies. This occurs because DG generates contexts about entities that remain within the LLM’s knowledge distribution, allowing parametric knowledge to override the generated evidence. In contrast, Figure~\ref{fig:quality_perturbation}(b) shows that SeedRG achieves 1--14\% leakage and 0\% factual inconsistencies. Additional results are provided in Appendix Table~\ref{tab:dg_failures}.

%Figure~\ref{fig:quality_perturbation}(a) shows DG is problematic: only 0--11\% of its questions are non-leaking. The rest either leak (19--39\%) or hallucinate (53--70\%). This happens because DG fabricates contexts about real entities the LLM already knows about, which makes the generation still falls in \textbf{LLM's knowledge distribution} Figure~\ref{fig:quality_perturbation}(b) shows SeedRG has 1--14\% leakage and 0\% hallucination. Full results could be checked in Appendix Table~\ref{tab:dg_failures}.

\paragraph{Graph structure preservation.} %Figure~\ref{fig:quality_perturbation}(c) compares graph metrics (Nodes, Edges, Density, AvgDeg) against the original benchmark. SeedRG deviates by less than 5\% on all metrics, while DG deviates by up to 27\% on nodes and 38\% on edges. This structural divergence means DG uncontrollably changes reasoning difficulty.
Figure~\ref{fig:quality_perturbation}(c) compares graph statistics (number of nodes, edges, density, and average degree) against the original benchmark. SeedRG deviates by less than 5\% across all metrics, while DG deviates by up to 27\% in nodes and 38\% in edges. These structural shifts indicate that DG alters the underlying reasoning structure, leading to uncontrolled changes in difficulty.

\subsubsection{From Graph Structure to Reasoning Difficulty}
\label{sec:graph_structure_exp}

%To verify the link between graph structure and reasoning difficulty, we regenerate SeedRG contexts from graph triplets under two settings: preserving the original structure, and applying cyclic permutation to rewire edges while keeping the same entities. As shown in Figure~\ref{fig:quality_perturbation}(d), the no-structure-change condition nearly overlaps with SeedRG across all engines and datasets, while the structure-change condition drops. This confirms that reasoning difficulty is governed by graph structure, further proving the necessity of reasoning graph check in SeedRG pipeline.

To validate the connection between graph structure and reasoning difficulty, we regenerate SeedRG contexts from graph triplets under two settings: (1) preserving the original structure, and (2) applying cyclic permutations to rewire edges while keeping the same entities. 

\begin{figure*}[t]
\centering
\includegraphics[width=0.9\linewidth]{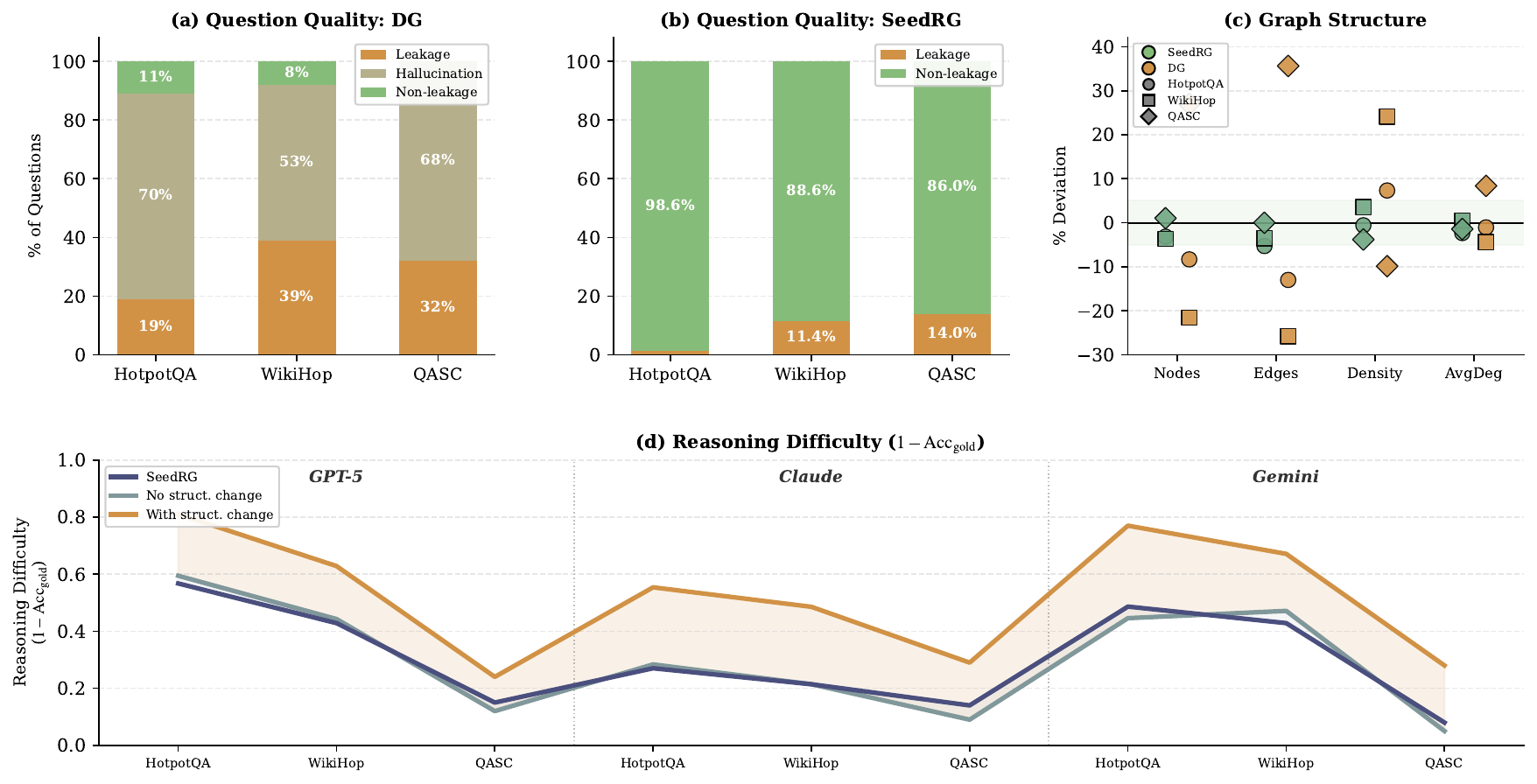}
\caption{Generation quality comparison: SeedRG vs Direct Generation. SeedRG produces non-leaking, hallucination-free questions while preserving graph structure, which leads to same reasoning difficulty.}
\label{fig:quality_perturbation}
\end{figure*}

As shown in Figure~\ref{fig:quality_perturbation}(d), the structure-preserving condition closely matches SeedRG across all models and datasets, while the structure-perturbed condition exhibits consistent performance degradation. This demonstrates that reasoning difficulty is governed by graph structure, and validates the necessity of the reasoning graph check in the SeedRG pipeline.

% ============================================================
% SECTION 4: RAG System Evaluation
% ============================================================
\subsection{RAG Algorithms Performance Comparison on SeedRG}
\label{sec:rag_eval}

%Having established that SeedRG satisfies leakage error, answerability accuracy, and difficulty preservation, we now use it to evaluate RAG algorithms. Figure~\ref{fig:rag_results} reports accuracy across all seven retrieval conditions on the SeedRG benchmark, using three QA engines. Full per-engine breakdowns are provided in Appendix Tables~\ref{tab:full_gpt5}--\ref{tab:full_gemini}.

\begin{figure*}[t]
\centering
\begin{subfigure}[t]{0.62\textwidth}
    \centering
    \includegraphics[width=\textwidth]{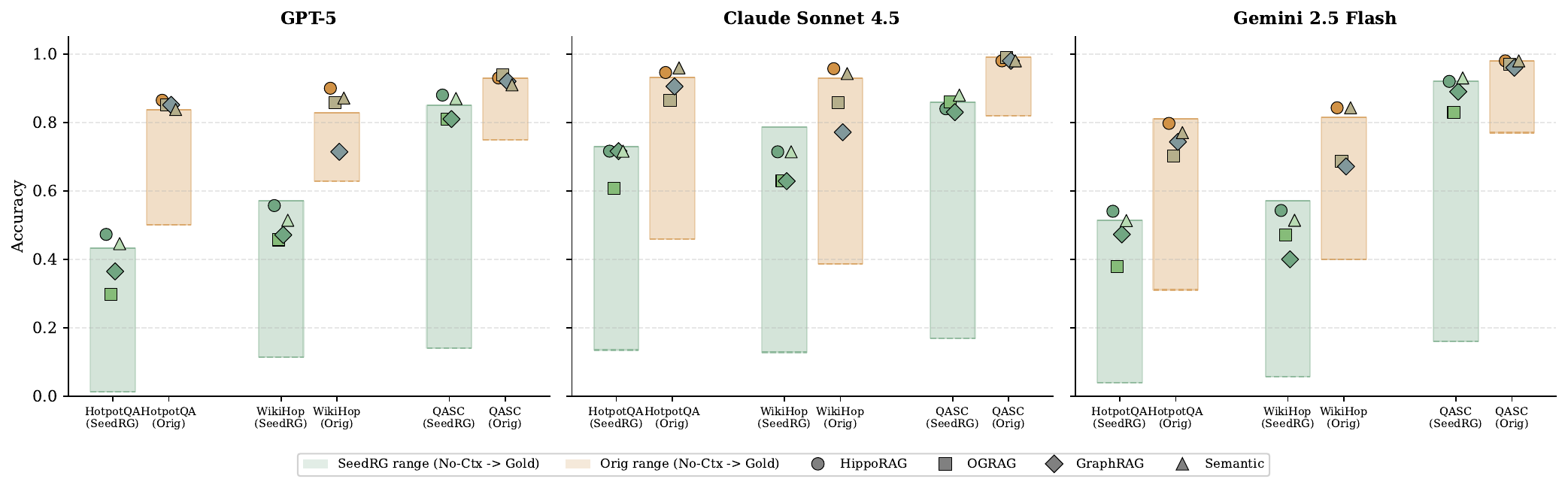}
    \caption{RAG algorithm evaluation: SeedRG vs Original benchmarks. SeedRG provides meaningful differentiation, while Original benchmarks cluster due to knowledge leakage.}
    \label{fig:rag_results}
\end{subfigure}
\hfill
\begin{subfigure}[t]{0.35\textwidth}
    \centering
    \includegraphics[width=\textwidth]{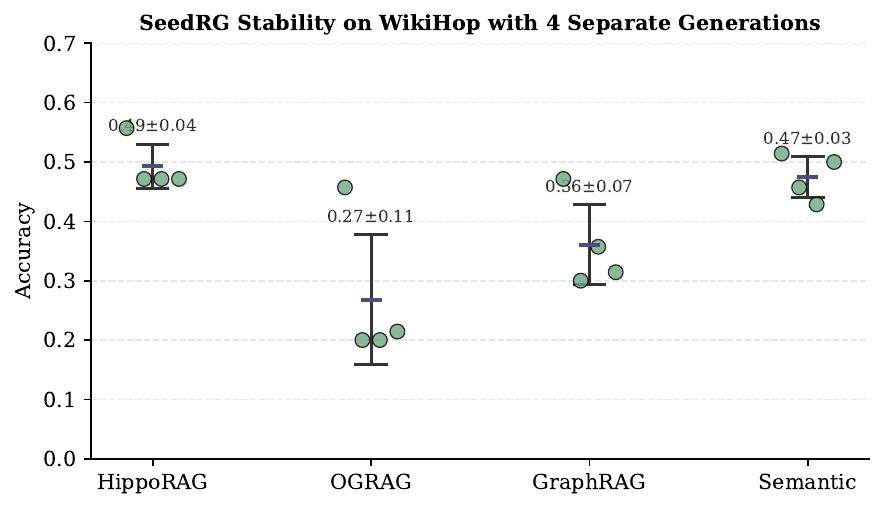}
    \caption{Benchmark stability: 4 independent SeedRG generations on WikiHop. SeedRG shows a stable winner/loser.}
    \label{fig:variability}
\end{subfigure}
\caption{RAG algorithm evaluation and stability on SeedRG.}
\label{fig:rag_eval}
\end{figure*}

Having established that SeedRG minimizes knowledge leakage, preserves answerability accuracy, and maintains reasoning difficulty, we now use it to evaluate RAG systems. Figure~\ref{fig:rag_results} reports accuracy across all retrieval conditions on SeedRG, using three QA engines. Full per-engine results are provided in Appendix Tables~\ref{tab:full_gpt5}--\ref{tab:full_gemini}.

On the original benchmarks, RAG systems show limited differentiation---on HotpotQA, all systems cluster within 3\% accuracy, making it impossible to distinguish retrieval quality. In contrast, SeedRG widens the spread to 10--18\% on HotpotQA and WikiHop, revealing genuine differences in retrieval effectiveness. For example, HippoRAG consistently outperforms OGRAG and GraphRAG across all engines, a ranking that is not observable on the original benchmarks. QASC exhibits a narrower range (.81--.93) due to its multiple-choice format, but still shows consistent differences across systems.

We also analyze the stability of benchmarks in Figure~\ref{fig:variability}. We take WikiHop benchmarks for example, and regenerate SeedRG 4 times. We still do the generation with GPT-4o-mini and evaluate each with GPT-5. As shown in Figure~\ref{fig:variability}, the standard deviation across runs is stable for all RAG systems.
\section{Conclusion and Future Work}

%We identify knowledge leakage as a critical problem in current multi-hop QA benchmarks: LLMs can answer 31--78\% of questions without any retrieval, making it impossible to reliably evaluate RAG algorithms. To address this, we propose SeedRG, a benchmark generation pipeline that automatically produces near-zero leakage, answerable, and difficulty-preserving benchmarks from any seed dataset. We further establish a causal link between reasoning graph structure and question difficulty, justifying the reasoning graph check as a core component of the pipeline.

%Our work currently treats difficulty preservation as a binary constraint. An open question is how to move from preservation to controlled difficulty change. Our future work would continually explore how to generate benchmarks at targeted difficulty levels.

We identify knowledge leakage as a fundamental failure mode in current multi-hop QA benchmarks. LLMs can answer 31--78\% of questions without retrieval, collapsing evaluation signal and obscuring differences between RAG systems. This issue compounds over time through benchmark aging, as benchmarks are absorbed into model training.

We propose \textbf{SeedRG}, a semi-synthetic pipeline that generates leakage-free, difficulty-preserving benchmarks from existing datasets. By preserving reasoning structure while replacing entities outside the model’s parametric knowledge, SeedRG enforces retrieval dependence and enables renewable benchmark generation. Empirically, SeedRG reduces leakage and restores discriminative power, revealing performance differences across RAG systems that are invisible on existing benchmarks.

Finally, we show that reasoning difficulty is governed by graph structure, providing a principled basis for controlling task difficulty. An important direction for future work is to move from preservation to control---enabling generation of benchmarks at targeted difficulty levels.
\newpage
\bibliography{colm2024_conference}
\bibliographystyle{colm2024_conference}

\appendix
\section{Additional Results}

\subsection{Generation Quality Examples}

Table~\ref{tab:dg_failures} shows concrete examples of the two failure modes discussed in Section~\ref{sec:quality_analysis}. In Example 1 (knowledge leakage), given the same seed question about a band's nationality, SeedRG replaces the band with a novel entity (``Mellow Vibes Harmony'') that the LLM cannot recognize, while DG generates a question about the well-known film ``Pan's Labyrinth''---the LLM immediately answers ``Mexican'' without needing any context. In Example 2 (factual hallucination), DG fabricates a band called ``The Echoes'' with lead guitarist ``Alex Chen,'' but the LLM's parametric knowledge associates ``The Echoes'' with a real band whose guitarist is ``Vic Briggs''---the generated context directly contradicts what the LLM knows. SeedRG avoids this by reusing original factual structures with only entity names changed.

\begin{table*}[h]
\centering
\footnotesize
\caption{Generation quality examples. Example 1 (WikiHop seed): DG produces a question about a well-known entity, causing knowledge leakage. Example 2 (HotpotQA seed): DG fabricates facts that contradict LLM knowledge. \textcolor{red}{Red} highlights critical differences.}
\label{tab:dg_failures}
\setlength{\tabcolsep}{4pt}
\begin{tabular}{l l p{4.2cm} p{2.2cm} p{2.2cm}}
\toprule
& & \textbf{Question} & \textbf{No-Context Answer} & \textbf{Ground Truth} \\
\midrule
\multicolumn{5}{l}{Example 1: Knowledge Leakage (WikiHop seed)} \\
\midrule
WikiHop   & & Lead singer of Hurlingham Reggae Band's nationality? & Italian--Scottish & Italian--Scottish \\
SeedRG & & Lead singer of \textcolor{red}{Mellow Vibes Harmony}'s nationality? & Unknown. & Galician--Welsh \\
DG     & & Director of \textcolor{red}{Pan's Labyrinth}'s nationality? & Mexican. & Mexican \\
\midrule
\multicolumn{5}{l}{Example 2: Factual Hallucination (HotpotQA seed)} \\
\midrule
HotpotQA   & & Lowest vocal range in Cosmos? & J\={a}nis Strazdi & J\={a}nis Strazdi\c{n}\v{s} \\
SeedRG & & Lowest vocal range in \textcolor{red}{Aetherius}? & Unclear.  & Raimonds B\={e}rzi\c{n}\v{s} \\
DG     & & Lead guitarist in \textcolor{red}{The Echoes}? & Vic Briggs. & Alex Chen \\
\bottomrule
\end{tabular}
\end{table*}

\subsection{Full Results: Entity Replacement vs.\ Direct Generation}
\label{app:full_results}

Tables~\ref{tab:full_gpt5}--\ref{tab:full_gemini} present the complete accuracy results across all RAG conditions, datasets, and generation methods. ``SeedRG'' denotes our pipeline; ``DG'' denotes Direct Generation; ``Orig'' denotes the original (unmodified) benchmark. SeedRG and Orig use N=74/70/100 questions for HotpotQA/WikiHop/QASC respectively; DG uses the same N with newly generated questions.

\begin{table*}[h]
\centering
\small
\setlength{\tabcolsep}{3pt}
\caption{Full results: GPT-5. Bold indicates the lowest no-context accuracy (least knowledge leakage).}
\label{tab:full_gpt5}
\begin{tabular}{l ccc ccc ccc}
\toprule
& \multicolumn{3}{c}{\textbf{HotpotQA}} & \multicolumn{3}{c}{\textbf{WikiHop}} & \multicolumn{3}{c}{\textbf{QASC}} \\
\cmidrule(lr){2-4} \cmidrule(lr){5-7} \cmidrule(lr){8-10}
\textbf{RAG} & SeedRG & DG & Orig & SeedRG & DG & Orig & SeedRG & DG & Orig \\
\midrule
No-Ctx   & \textbf{.014} & .190 & .500 & \textbf{.114} & .390 & .629 & \textbf{.140} & .320 & .750 \\
Gold     & .432 & .690 & .838 & .571 & .720 & .829 & .850 & .730 & .930 \\
HippoRAG & .473 & .660 & .865 & .557 & .690 & .900 & .880 & .710 & .930 \\
OGRAG    & .297 & .260 & .851 & .457 & .450 & .857 & .810 & .390 & .940 \\
GraphRAG & .365 & .240 & .851 & .471 & .440 & .714 & .810 & .460 & .920 \\
Semantic & .446 & .670 & .838 & .514 & .700 & .871 & .870 & .720 & .910 \\
\bottomrule
\end{tabular}
\end{table*}

\begin{table*}[h]
\centering
\small
\setlength{\tabcolsep}{3pt}
\caption{Full results: Claude Sonnet 4.5.}

\label{tab:full_claude}
\begin{tabular}{l ccc ccc ccc}
\toprule
& \multicolumn{3}{c}{\textbf{HotpotQA}} & \multicolumn{3}{c}{\textbf{WikiHop}} & \multicolumn{3}{c}{\textbf{QASC}} \\
\cmidrule(lr){2-4} \cmidrule(lr){5-7} \cmidrule(lr){8-10}
\textbf{RAG} & SeedRG & DG & Orig & SeedRG & DG & Orig & SeedRG & DG & Orig \\
\midrule
No-Ctx   & \textbf{.135} & .310 & .459 & \textbf{.129} & .420 & .386 & \textbf{.170} & .530 & .820 \\
Gold     & .730 & .780 & .932 & .786 & .880 & .929 & .860 & .820 & .990 \\
HippoRAG & .716 & .790 & .946 & .714 & .820 & .957 & .840 & .860 & .980 \\
OGRAG    & .608 & .330 & .865 & .629 & .430 & .857 & .860 & .680 & .990 \\
GraphRAG & .716 & .330 & .905 & .629 & .470 & .771 & .830 & .650 & .980 \\
Semantic & .716 & .810 & .959 & .714 & .830 & .943 & .880 & .850 & .980 \\
\bottomrule
\end{tabular}
\end{table*}

\begin{table*}[h]
\centering
\small
\setlength{\tabcolsep}{3pt}
\caption{Full results: Gemini 2.5 Flash.}
\label{tab:full_gemini}
\begin{tabular}{l ccc ccc ccc}
\toprule
& \multicolumn{3}{c}{\textbf{HotpotQA}} & \multicolumn{3}{c}{\textbf{WikiHop}} & \multicolumn{3}{c}{\textbf{QASC}} \\
\cmidrule(lr){2-4} \cmidrule(lr){5-7} \cmidrule(lr){8-10}
\textbf{RAG} & SeedRG & DG & Orig & SeedRG & DG & Orig & SeedRG & DG & Orig \\
\midrule
No-Ctx   & \textbf{.041} & .180 & .311 & \textbf{.057} & .350 & .400 & \textbf{.160} & .380 & .770 \\
Gold     & .514 & .640 & .811 & .571 & .730 & .814 & .920 & .760 & .980 \\
HippoRAG & .541 & .610 & .797 & .543 & .730 & .843 & .920 & .770 & .980 \\
OGRAG    & .378 & .070 & .703 & .471 & .230 & .686 & .830 & .470 & .970 \\
GraphRAG & .473 & .190 & .743 & .400 & .350 & .671 & .890 & .450 & .960 \\
Semantic & .514 & .620 & .770 & .514 & .710 & .843 & .930 & .790 & .980 \\
\bottomrule
\end{tabular}
\end{table*}

\end{document}